# Optimal Motion of Flexible Objects with Oscillations Elimination at the Final Point

**Natalia Varminska and Damien Chablat**

[1] *Sevastopol State University, Russian, e-mail: nvarminska@gmail.com*

[2] *IRCCyN, CNRS, France e-mail: Damien.Chablat@cnrs.fr*

**Abstract**. In this article, a theoretical justification of one type of skew-symmetric optimal translational motion (moving in the minimal acceptable time) of a flexible object carried by a robot from its initial to its final position of absolute quiescence with the exception of the oscillations at the end of the motion is presented. The Hamilton-Ostrogradsky principle is used as a criterion for searching an optimal control. The data of experimental verification of the control are presented using the Orthoglide robot for translational motions and several masses were attached to a flexible beam.

## 1.    Introduction

Considering the flexible object moving requirements and the robotic industrial needs, the most important one is the evaluation and the possible suppression of the robots oscillations or the oscillations of the objects which the robots move. Actually, robots with reduced structure masses are appeared and as a result they lose the structural rigidity thus affecting the system dynamics. Use of structures and materials that can suppress the oscillations is quite expensive and furthermore not fully acceptable [1, 2]. Additional approaches based on software solutions need a calibration of the control loop and a specific path planning phase to decrease the positioning errors [2 – 4].

The trajectory planning of a robot generally means developing a mathematical algorithm for the selection and the description of the desired motion between the initial and the final points of the trajectory. To do this, two approaches are usually used [5 – 8]. In the first one, the exact set of constraints (for example, continuity and smoothness of the functions) on the position, velocity and acceleration of the generalized coordinates of the robot in some points of the trajectory are specified. Then the planner selects the trajectory that passes through the needed points and satisfies the given constraints into them from a class of functions (for example, among the polynomials). The trajectory developed by the planner must take into consideration the mechanical constraints of the robot and, hence, a smooth trajectory has to be searched. To satisfy this demand, it would be desirable to get the trajectories with continuous functions of the joint accelerations, so that the jerk stays limited. The restrictions definition and planning of the trajectory are performed in the joint space. In the second one, the desired trajectory of the robot is described as an analytic function, for example, a linear trajectory in Cartesian workspace. The planner makes an approximation of the desired path in Cartesian or joint coordinates. The trajectory planning in joint variables has several advantages: a) there is the definition of the variables directly controlled during the motion; b) trajectory planning can be performed in real-time; c) trajectory planning in joint coordinates is easier to execute. The drawback is the complexity to determine the position of robot links and the end-effector during the motion. The same tendency can be noted for moving flexible objects. Trajectory planning techniques aim at the minimization of some objective functions that usually are the implementation time, the actuator efforts and the jerk [6, 9, 10, 12]. Due to the need to increase productivity in industry, the first used trajectory planning technique is the minimum time algorithm [11 – 14]. The main drawback of these approaches is that the generated trajectories have discontinuous accelerations and the joint torque values and they bring in the dynamic difficulties in the trajectory implementation. Some robot trajectory planning algorithms with power criterions are employed, for example, in [6, 7, 12, 15].

The next algorithm provides a jerk-optimal trajectory by minimizing the jerk value during the trajectory performance. According to the algorithm, the positioning errors, the actuators stresses and the robot structures (i.e., definition of the resonance frequencies) can be detected in [16, 17]. While planning the trajectory of a flexible object, a manipulator can move with the object slowly enough to respond to the velocity and torque constraints [18 – 20]. In these papers, the motion is divided into three phases: a motion with acceleration, a motion with constant velocity and the deceleration. The first phase lasts until the velocity boundary is achieved, then the constant-velocity phase starts. In proposed method the torque, dynamic and kinematic constraints are satisfied and the needed object position and velocity can be reached at the end point. In [18], it was presented a trajectory planning approach that satisfies the dynamic constraints (the torque and the velocity constraints) and in this way the position and the velocity conditions at the final point could be satisfied both. In the paper the authors proposed to set apart the task of planning the direction of a robot and the task of the speed-planning. There are studies on the control of oscillations of linear and nonlinear mechanical systems in absolute motion [21, 22]. In [23, 24], the authors focus on optimal control of translational and rotational motions of the flexible systems with finite or infinite number of degrees of freedom. When transporting in a minimal time of non-rigid objects or moving a robot with limited rigidity, the oscillations of the robot and these transported objects occur. But the dynamics of non-rigid robots in optimal translational motion deserves special attention. Research of the optimal motion control of the flexible objects with the elimination of oscillations at the end of the motion is required. The control tasks of flexible systems are relevant in using robots with finite rigidity (robots of minimal mass), transportation and assembly of the flexible objects under terrestrial conditions and in outer space. There is a need to use such special motion controls, in which oscillations of transported objects are significantly reduced or completely eliminated, i.e. in an acceptable minimum possible time of translational motion the relative or absolute quiescence at the end of the movement is achieved [25, 26]. The proposed motion control provides moving a flexible object (or a non-rigid robot arm) from the initial position of absolute quiescence to the end position of absolute quiescence with oscillation acceptability only during the motion and elimination of them at the end point. So there is no positioning accuracy loss. The functions describing the positions, the velocities and the accelerations in translational motion are smooth and they are functions of the object natural frequency and motion time. The total motion time also depends on the object natural frequency.

This paper is composed in the following way. The first section gives theoretical justification of an optimal translational motion using the Hamilton-Ostrogradsky principle. The next section deals with the experimental verification of this motion. Here the results of practical investigation of the flexible object motions with different masses are presented. Finally, the conclusions are made and a future research is presented.

## 2. Theoretical justification of an optimal translational motion

A flexible object participates in two motions: a translational or rotational motion, (this is a motion with the moving coordinate system) and the relative motion (this motion is relative to moving coordinate system, i.e. oscillations). The Hamilton-Ostrogradsky principle can be used as a criterion for searching an optimal control of translational motions of flexible objects [27]. This motion can be optimal in the sense of some optimality criterion (which can be known before or should be found as a result of research). The possibility of using the Hamilton-Ostrogradsky principle (in the Lagrange form) as a criterion for the motion optimality of the rigidly-flexible frames employing a pulse force is considered in [28]. In [29] it is shown that the dry friction in studying the natural oscillations can be considered as a relay control which is found using the Pontryagin maximum principle on the basis of the time criterion or the action principle.

For justification of the optimal control, the Hamilton-Ostrogradsky principle is involved and it means that for the non-conservative (controlled) system,

$$J = \int_0^{t_1} \left( T - \Pi + A^e \right) dt, \qquad (1)$$

the action takes a stationary value in real (true) motion. For example, for a system with one degree of freedom the kinetic energy of the translational motion at any moment of time is $T = m v_e^2 / 2$, where $m$ is the mass of an object; $v_e$ is the translational velocity at any moment of time. The potential energy stored during the optimal fast translational motion of the flexible object is $\Pi = c s_e^2 / 2 n^2$, where $c$ is the rigidity coefficient; $s_e$ is the translational displacement at any moment of time; $n = t_1 / t_c$, where $t_1$ is the total motion time; $t_c$ is the period of the natural oscillations of an object (in relative motion). Essentially, using $n$ takes into account the deformation of the flexible object due to the translational motion. The higher the natural frequency $k$ of oscillation of an object $(t_c = 2\pi / k)$, i.e. the higher rigidity of the flexible connection, the less potential energy of deformation stored as a result of the fairly rapid translational motion.

The mechanical work of the control force at any moment of time is $A^e = u_e^* s_e$, where

- $u_e^* = m s_e^* p^2$, and $s_e^* = V_{cp} t = L t / t_1$; $p = k / n$;
- $L$ is the overall displacement of an object during the motion $t_1$;
- $V_{cp} = L / t_1$ is average speed.

After substituting in Eq. (1), the criterion takes the form:

$$J = \int_0^{t_1} \left( \frac{m v_e^2}{2} - \frac{c s_e^2}{2 n^2} + \frac{m L p^3}{2\pi} t s_e \right) dt . \qquad (2)$$

For the functional (2) that for this case in the general form is written as

$$\int_0^{t_1} F(s_e, \dot{s}_e, t) dt,$$

the Euler equation is

$$F_{s_e} - \frac{d}{dt} F_{\dot{s}_e} = 0 .$$

After transformations for $m = 1$, we receive

$$\frac{d^2 s_e}{dt^2} + p^2 s_e = \frac{L p^3}{2\pi} t. \qquad (3)$$

The solution of Eq. (3) with the initial conditions $s_e(0) = 0$ and $\dot{s}_e(0) = v_e(0) = 0$ is:

$$s_e(t) = \frac{L}{2\pi} \left( p t - \sin(p t) \right),$$

and then the velocity and acceleration are:

$$v_e(t) = \dot{s}_e(t) = \frac{L p}{2\pi} \left( 1 - \cos(p t) \right); \quad u_e(t) = \dot{v}_e(t) = \frac{L p^2}{2\pi} \sin(p t) \qquad (4)$$

Functions $s_e(t)$, $v_e(t)$ and $u_e(t)$ for initial data: $L = 0.41$ m, $k = 5.78$ s$^{-1}$ and $p = k / n$ are depicted in Fig. 1.

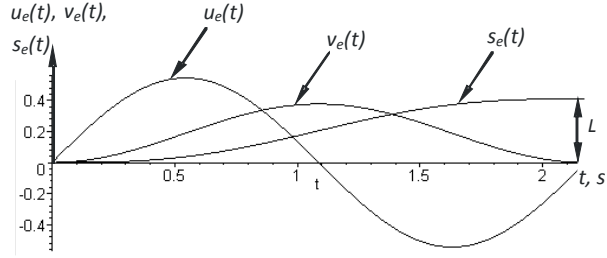

Figure 1. **Translational motion graphs: $s_e(t)$ displacement, $v_e(t)$ velocity, $u_e(t)$ acceleration**

They satisfy the boundary conditions of the translational motion $u_e(0)=0$, $u_e(t_1)=0$, $v_e(0)=0$, $s_e(0)=0$ and the following additional conditions at the right end, which can be represented as the moment ratios:

$$\int_0^{t_1} u_e(t)dt = 0; \qquad \int_0^{t_1} v_e(t)dt = L.$$

The oscillator differential equation in the relative motion (oscillations) is

$$\ddot{x}_r(t) + k^2 x_r(t) = -\frac{Lp^2}{2\pi}\sin(pt)$$

and its solution with zero initial conditions $\left(x_r(0)=0,\quad \dot{x}_r(0)=0\right)$ is as follows,

$$x_r(t) = \frac{Lp^2}{2\pi(k^2 - p^2)}\left(\frac{p}{k}\sin(kt) - \sin(pt)\right).$$

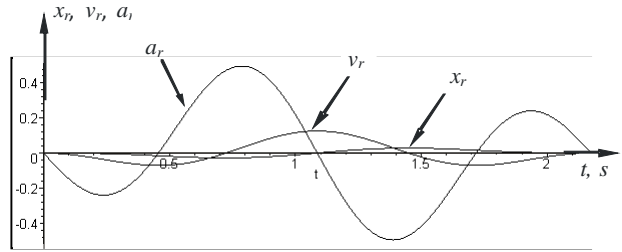

Figure 2. **Relative motion graphs: $x_r$ displacement, $v_r$ velocity, $a_r$ acceleration**

Then the velocity and acceleration in relative motion are:

$$v_r(t) = \frac{Lp^3}{2\pi(k^2 - p^2)}\left(\cos(kt) - \cos(pt)\right), \quad a_r(t) = \frac{Lp^3}{2\pi(k^2 - p^2)}\left(p\sin(pt) - k\sin(kt)\right)$$

Figures 1 and 2 show that at the end point of the displacement, velocity and acceleration in relative motion are equal zero. Thus the velocity and acceleration in translational motion is equal to zero and the displacement is equals to *L*. So there is the position of absolute quiescence with needed displacement. The moment ra-tios at the right end, which mean that relative displacement and velocity equal zero at $t = t_1$, are written

$$\int_0^{t_1} u_e(t)\cos(kt)dt = 0; \qquad \int_0^{t_1} u_e(t)\sin(kt)dt = 0. \qquad (5)$$

The relations (5) with accounting (4) and $p = k/n$, $t_1 = 2\pi/p$ are transformed in transcendental equations:

$$\cos(2n\pi) - 1 = 0, \qquad \sin(2n\pi) = 0. \qquad (6)$$

Equation (6) with the exception of the resonance has a solution: $n = 2, 3, \ldots$ . Found $n$ is used to calculate the total motion time $t_1$. It should be noted that it could be found $n$ for which during the time $t_1$ a flexible object moves from the initial to the final state for all skew-symmetric controls for which is true

$$\int_0^{t_1} u_e(t)dt = 0, \; u_e(0) = 0 \text{ and } u_e(t_1) = 0.$$

## 3. Experimental verification of the optimal translational motion

So, there was made an experiment of the optimal translational motion of a flexible beam from an initial point to the final point in a minimal time that is consistent with the natural frequency of the beam. The experiment was made in the IRCCyN's laboratory using the Orthoglide 5-axis robot (Figs 3 a) [30, 31] built by Symetrie (France) and following equipment: an accelerometer of series FA 101 (FGP Instrumentation) and DS1103 PPC Controller Board. The trajectories of the robot were calculated using Matlab and supplied to the DS PPC Controller Board card. The actuator positions are acquired with a frequency equal to 9 kHz but the robot motions are controlled thanks to a sub program working at 1.5 kHz. As the position information transits the drives which perform the encoder emulation. Noise on the position and speed appears even when the robot is stopped. The amplitude of this noise is about four micrometers.

The oscillator's base is clamped in the chuck of milling of the Orthoglide 5-axis robot. It moves under the influence of the control function $u_e(t) = a\sin(pt)$ with time agreed with the first period of the oscillations of the flexible system (Fig. 3b). This system is a beam with a rectangular cross-section (with $l= 0.305$ m, $b= 0.013$ m, $h= 0.5 \cdot 10^{-3}$ m and Young's modulus $E= 2.1 \cdot 10^{11}$ Pa) and with a mass located at its tip. The graphs of the oscillations of the object obtained with the accelerometer (for initial data: $L= 0.41$ m, $m= 0.09$ kg and $k= 5.78$ s$^{-1}$) are given in the Figs. 4 without filtering and 5 with filters. Filtering is needed to cut off high pass frequencies (noise) in the signal which are produced by the accelerometer. For this purpose, here, a Butterworth filter is used. This is a type of signal processing filter organized to get a frequency response as flat as possible in the passband. The Butterworth filter rolls off more slowly around the cutoff frequency than the Chebyshev filter or the Elliptic filter but it has no ripple. And zero-phase filter helps to save features in a filtered time waveform exactly where they occur in the unfiltered signal.

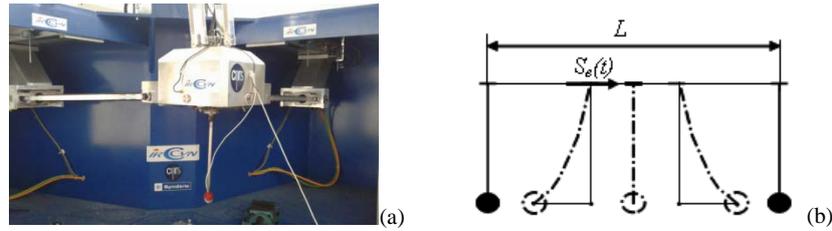

Figure 3. **a) scheme of the experiment using semi-industrial prototype of the Orthoglide 5-axis; b) scheme of the movement of an object with one degree of freedom**

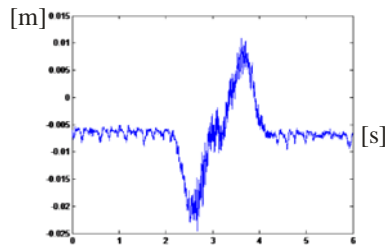 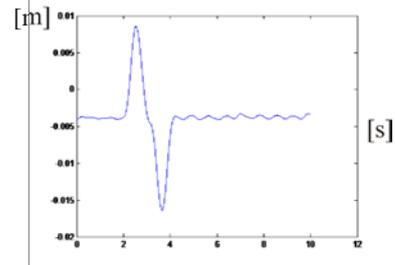

Figure 4. **Oscillations of the object tip for $m=$ 0.09 kg without filtering**      Figure 5. **Oscillations of the object tip for $m=$ 0.09 kg with filtering**

For the experiments, a 4th order Butterworth filter was used with the following parameters of the filtering: the band pass filter is [0 : 20]Hz and hence the cut-off frequency is 20 Hz. The graphs confirm the absence of the oscillations at the end of the motion but in the graphs it could be noticed a ''measurement noise'' before and after measuring. If the motion time is calculated regardless to the first period of the object oscillation, there are significant displacements at the end of the motion even if the motion time is bigger (Fig. 6). For this case, the amplitude is $\pm\, 3 \div 4$ mm. There were obtained the oscillation graphs for a mass $m=$ 0.06 kg (Fig. 7 and 8). Here, it could be seen that the oscillations are significantly decreased but not eliminated at all. It can be explained by the inaccuracies in initial data of experiment. But elimination of the oscillations could be reached by using control feedback.

In the first case, the amplitude of the oscillations (with regard to the natural period of the oscillations) is $\pm$1 mm at the object tip and, in the second case (regardless to the object oscillations period), the amplitude is $\pm\, 6 \div 8$ mm. So, for this one, using the proposed motion control allows to decrease the oscillations almost 6 times.

The experiments are also made with the following masses 0.02 kg, 0.075 kg and 0.15 kg. In all cases, we observe the same behavior of the system. The results are presented in the table 1.

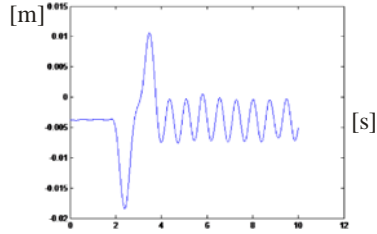

Figure 6.  **Oscillations of the object tip for $m$= 0.09 kg**

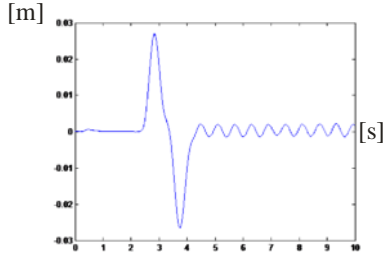 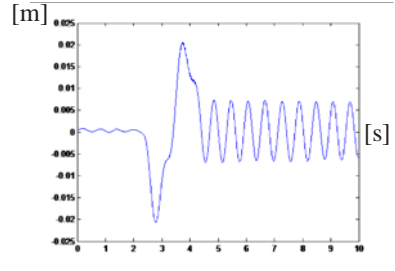

Figure 7.  **Oscillations of the object tip for $m = 0.06$ kg**

Figure 8.  **Oscillations of the object tip for $m = 0,06$ kg regardless to its first period**

TABLE I.  AMPLITUDE OF THE OSCILLATIONS OF THE OBJECT TIP

| $m$ [kg] | 0.02 | 0.06 | 0.075 | 0.09 |
|---|---|---|---|---|
| Amplitude regardless to the natural period of the object oscillations [mm] | ± 7 ÷ 9 | ± 6 ÷ 8 | ± 5 ÷ 6 | ± 3 ÷ 4 |
| Amplitude with regard to the object oscillations period [mm] | ± 2 ÷ 3 | ± 1 | ± 0.5 | 0.0 |

The energy expenditure for the operations performance was estimated. For example, for an object mass $m$= 0.09 kg changing the motion time by 5 % produces the 12 ÷ 18 % energy drop. For an object mass $m$ = 0.06 kg changing the motion time by 10 % entails the energy costs by 60 ÷ 70 %.

## 4.  Conclusions

It was proved during the experiment that only increasing the motion time does not remove the oscillations of the flexible system but use of the proposed control and choosing the motion time depending on an object natural frequency eliminate the oscillations at the end point. Nowadays, the implementation of the necessary precise time for the technological operations of the object movements is a simple task for the majority of industrial robots. So it is convenient to use the proposed motion control, as it saves energy $1.5 \div 2$ times (depending on the object mass). The suggested control may be effectively used to eliminate the oscillations of the flexible systems or to decrease them $3 \div 10$ times (depending on the object mass and motion time) during the transporting operations, the assembly of flexible objects under terrestrial conditions or in outer space. In future works, it is planned to take into account the dry and the linear-viscous resistances in the object motions and investigate the motions from non-zero initial conditions.

**Acknowledgements** The work presented in this paper was partially funded by the Erasmus Mundus project "Active". Both authors also thank A. Jubien, E. Besnier and P. Lemoine for their technical assistance during the experiments.